\documentclass{article}

\usepackage[preprint]{neurips_2024}

\usepackage[utf8]{inputenc} % allow utf-8 input
\usepackage[T1]{fontenc}    % use 8-bit T1 fonts
\usepackage{hyperref}       % hyperlinks
\usepackage{url}            % simple URL typesetting
\usepackage{booktabs}       % professional-quality tables
\usepackage{amsfonts}       % blackboard math symbols
\usepackage{amsmath}   % For math environments and symbols
\usepackage{amssymb}   % For \mathbb
\usepackage{amsthm}    % For argmin/argmax

\usepackage[ruled,vlined]{algorithm2e}

\usepackage[utf8]{inputenc}
\usepackage{nicefrac}       % compact symbols for 1/2, etc.
\usepackage{microtype}      % microtypography
\usepackage{xcolor}         % colors
\usepackage{graphicx}

\DeclareMathOperator*{\argmin}{arg\,min}
\DeclareMathOperator*{\argmax}{arg\,max}
\DeclareMathOperator{\E}{\mathbb{E}} % For Expectation
\DeclareMathOperator{\DJS}{D_{JS}} % For JS-Divergence
\newcommand{\BE}{B_E}
\newcommand{\BBeta}{B_\beta}

\newcommand{\Ptarget}{P'_{\text{target}}}

\title{Enabling Off-Policy Imitation Learning with Deep Actor-Critic Stabilization}

\author{%
  Sayambhu Sen \\
  Amazon Alexa\\
  \texttt{sensayam@amazon.com} \\
  \And
  Shalabh Bhatnagar\\
  Indian Institute of Science \\
  \texttt{shalabh@iisc.ac.in} \\
}

\begin{document}

\maketitle

\begin{abstract}
  Learning complex policies with Reinforcement Learning (RL) is often hindered by instability and slow convergence, a problem exacerbated by the difficulty of reward engineering. Imitation Learning (IL) from expert demonstrations bypasses this reliance on rewards. However, state-of-the-art IL methods, exemplified by Generative Adversarial Imitation Learning (GAIL) \cite{ho2016gail}, suffer from severe sample inefficiency. This is a direct consequence of their foundational on-policy algorithms, such as TRPO \cite{schulman2017trustregionpolicyoptimization}. In this work, we introduce an adversarial imitation learning algorithm that incorporates off-policy learning to  improve sample efficiency. By combining an off-policy framework with auxiliary techniques, in this case a double Q network based stabilization and value learning without reward function inference, we demonstrate a reduction in the samples required to robustly match expert behavior.
\end{abstract}

\section{Introduction}
While deep reinforcement learning (RL) has achieved notable success in solving complex control tasks ( \cite{Mnih2015HumanlevelCT} ; \cite{schulman2017trustregionpolicyoptimization}; \cite{lillicrap2019continuouscontroldeepreinforcement}), its practical application is often hindered by the difficulty of reward engineering. The performance of RL algorithms is highly sensitive to the reward specification, and hand-crafting a function that elicits the correct behavior is a formidable challenge.

In many domains where designing a reward is difficult, providing expert demonstrations is comparatively straightforward. For example, it is simpler to demonstrate correct driving or robotic grasping than to formulate a reward function that perfectly encodes those skills. This paradigm motivates Imitation Learning (IL) \cite{Osa_2018_imitation}, which learns policies from demonstrations without a hand-engineered reward.

The limitations of reward-based learning are particularly stark in sparse-reward settings. Although specialized algorithms like Hindsight Experience Replay (HER) \cite{andrychowicz2018hindsightexperiencereplay} exist, they are often restricted to goal-oriented tasks. For complex behaviors like robotic locomotion, which may lack a clear goal structure but still suffer from sparse signals, demonstrations offer a richer, more effective source of supervision. Therefore, IL provides a powerful framework for tasks where reward engineering is a bottleneck or where rewards are sparsely available.

\section{Previous Methods}

\subsection{Behaviour Cloning (BC)}
Behaviour Cloning (BC) is the simplest method of imitation learning, framing it as a standard supervised learning problem. It learns a policy by treating state-action pairs $(s_t, a_t)$ from a dataset of expert demonstrations $\mathcal{D}_{\text{expert}}$ as a static collection of (input, target) pairs.

The goal is to learn a function of the form:
\begin{equation}
    a_t \sim \pi_\theta(\cdot | s_t)
\end{equation}
where we assume $\pi_\theta$ is a stochastic policy parameterized by $\theta$ (typically a neural network). The policy's output representation depends on the action space:
\begin{itemize}
    \item \textbf{Discrete Actions:} For a space with (say) $K$ actions, the policy network outputs a softmax distribution over the $K$ choices, given the state input $s_t$.
    \item \textbf{Continuous Actions:} The function approximator for the policy outputs the parameters of a distribution function (e.g., the mean and variance of a Gaussian). The action $a_t$ is then chosen by sampling from this distribution.
\end{itemize}

The training for the neural network is performed via stochastic gradient ascent to maximize the log-likelihood of the expert demonstrations w.r.t. the parameters $\theta$ \cite{Osa_2018_imitation}. The gradient is given by:
\begin{equation}
    \nabla_\theta J(\theta) \propto \frac{\partial \ln \pi_\theta(a_t|s_t)}{\partial \theta}
\end{equation}
where $\pi_\theta(a_t|s_t)$ is the probability density (or mass) of the expert action $a_t$ given the expert state $s_t$ under the policy $\pi_\theta$.

\subsubsection*{Limitations of Behaviour Cloning}
The primary limitation of BC, particularly when only expert trajectories are available, is \textbf{covariate shift}. This refers to the distributional mismatch that occurs when the training data distribution (from the expert) differs from the test data distribution (encountered by the learned policy).

This occurs mainly in the case when the number of expert trajectories is limited.
This leads to incomplete coverage of the expert distribution, and hence, the expert data distribution cannot be recovered. Because the policy is untrained on this new state distribution, its behavior can become undefined and erratic, leading to a compounding of errors and catastrophic failure.

This problem is particularly severe when the number of expert trajectories is limited, as the expert's state-visitation distribution cannot be fully recovered. In complex, high-dimensional state and action spaces, the number of trajectories required for complete coverage grows exponentially due to the \textbf{curse of dimensionality}. Given that collecting expert data is often costly, the dataset is almost always sparse. Consequently, Behaviour Cloning policies often fail to generalize and cannot produce robust, optimal policies.

\subsection{Inverse Reinforcement Learning (IRL)}

Inverse Reinforcement Learning (IRL) methods do not attempt to learn a direct mapping from states to expert actions. Instead, IRL seeks to recover the underlying \textbf{cost function} (or, equivalently, the reward function) that rationalizes the expert's behavior. The central hypothesis is that the expert policy $\pi_E$ is (near-)optimal with respect to this latent cost function. Once this function is learned, a standard reinforcement learning algorithm can be applied to find a policy that is similar to, or even better than, the expert's.

A primary challenge in IRL, especially with a limited number of demonstrations, is that the problem is \textbf{ill-posed}: many different cost functions can explain the same expert trajectories. To resolve this ambiguity, IRL methods typically impose additional constraints or priors to recover a plausible cost function.

For example, \cite{abbeelngapleirl2004} proposed to learn a policy by matching the \textbf{feature expectations} of the expert. The feature expectation $\mu(\pi)$ for a policy $\pi$ is defined as:
\begin{equation}
    \mu(\pi) = E \left[ \sum_{t=0}^{T} \gamma^t \phi(x_t) \Big| \pi \right] \in \mathbb{R}^k
\end{equation}
where $\phi(x_t)$ is a $k$-dimensional feature vector at state $x_t$. If the reward function is linear in these features, $R(x_t) = w^T \phi(x_t)$, the value of a policy is given by:
\begin{equation}
    E[R | \pi] = w^T \mu(\pi)
\end{equation}
The algorithm then finds a policy $\tilde{\pi}$ whose feature expectations are close to the expert's, $\mu(\tilde{\pi}) \approx \mu(\pi_E)$.

Another approach is \cite{maxmarginratliff2006}, which frames IRL as a max-margin structured prediction problem. MMP finds a cost function, parameterized by $w$, such that the cost of the demonstration trajectory $C_w(\tau_{\text{demo}})$ is lower than the cost of all other alternate trajectories $C_w(\tau)$ by a margin that scales with a loss function $L(\tau)$. A simplified constraint can be expressed as:
\begin{equation}
    C_w(\tau_{\text{demo}}) \le \min_{\tau} \{ C_w(\tau) - L(\tau) \}
\end{equation}
where $L(\tau)$ is the loss function.

A powerful and widely used constraint is the \textbf{principle of maximum entropy}. \cite{maxentirl2008ziebart} showed that by maximizing the entropy of the policy's trajectory distribution, subject to the constraint of matching feature counts, one can uniquely recover the optimal cost. This leads to a game-theoretic formulation where the IRL step finds the cost function that best distinguishes the expert from other policies:
\begin{equation}
    \text{IRL}(\pi_E) = \max_{c \in \mathcal{C}} \min_{\pi \in \Pi} \left[ -H(\pi) + E_{\pi}[c(s,a)] - E_{\pi_E}[c(s,a)] \right]
\end{equation}
where $H(\pi) = E_{\pi}[-\log \pi(a|s)]$ is the policy entropy.

\subsubsection*{Limitations of IRL}
A significant practical drawback of most IRL methods is their computational expense. They typically involve a "nested-loop" structure:
\begin{enumerate}
    \item \textbf{Outer-loop (IRL):} Update the estimate of the cost function $c(s,a)$.
    \item \textbf{Inner-loop (RL):} Solve for the optimal policy $\pi^*$ given the current cost $c(s,a)$. This step is often a full reinforcement learning process:
    \begin{equation}
        \text{RL}(c) = \argmin_{\pi \in \Pi} \left[ -H(\pi) + E_{\pi}[c(s,a)] \right]
    \end{equation}
\end{enumerate}
The RL step must be run to convergence within each iteration of the IRL step. This inner RL loop can require a large number of environment interactions, making the entire algorithm prohibitively slow and sample-inefficient.

\subsection{Generative Adversarial Imitation Learning (GAIL)}
\cite{ho2016gail} established a formal connection between Inverse Reinforcement Learning (IRL) and generative modeling. They demonstrated that the max-entropy IRL problem is the dual of an occupancy measure matching problem, and that the resulting optimal policy is the primal optimum.

Leveraging this insight, they proposed Generative Adversarial Imitation Learning (GAIL), an algorithm that frames imitation learning within a Generative Adversarial Network (GAN) \cite{goodfellow2014generativeadversarialnetworks} framework. The objective of GAIL is to learn a policy $\pi$ whose state-action occupancy measure, $\rho_\pi$, directly matches the expert's occupancy measure, $\rho_{\pi_E}$. This avoids the computationally expensive nested loop of traditional IRL.

The algorithm seeks to solve the following problem, regularized by the causal entropy $H(\pi)$ with a coefficient $\lambda \ge 0$:
\begin{equation}
    \min_{\pi} \max_{\psi} \psi^*_{\text{GA}}(\rho_\pi, \rho_{\pi_E}) - \lambda H(\pi)
\end{equation}
where $\psi^*_{\text{GA}}$ represents the GAN divergence objective. When using the standard binary cross-entropy loss from GANs, this objective is equivalent to minimizing the Jensen-Shannon divergence $D_{JS}$:
\begin{equation}
    \min_{\pi} D_{JS}(\rho_\pi, \rho_{\pi_E}) - \lambda H(\pi)
\end{equation}

This optimization is achieved by imposing a cost function regularizer that makes the occupancy measure matching loss nearly identical to the standard GAN loss. In practice, this is solved as a saddle-point problem by finding the minimum w.r.t. the policy $\pi$ and the maximum w.r.t. a discriminator $D$:
\begin{equation}
    \min_{\pi} \max_{D} \quad \mathbb{E}_{\pi}[\log(D(s,a))] + \mathbb{E}_{\pi_E}[\log(1-D(s,a))] - \lambda H(\pi)
\end{equation}

The GAIL algorithm is implemented using (at least) two neural networks:
\begin{enumerate}
    \item \textbf{A policy $\pi_\theta$} (the "generator" or "actor"), which maps states to actions.
    \item \textbf{A discriminator $D_\phi$} (the "critic"), which attempts to distinguish between expert and policy state-action pairs.
\end{enumerate}
If the policy is trained with an actor-critic RL algorithm, a third network for the value function (the "critic" in the RL sense) is also used.

The training alternates between two steps:
\begin{itemize}
    \item \textbf{Discriminator Update:} The discriminator $D_\phi$ is trained as a binary classifier using a supervised loss. State-action pairs from the expert trajectories $\mathcal{D}_{\text{expert}}$ are given a label of 1 (real), while pairs $(s, a) \sim \pi_\theta$ generated by the policy are given a label of 0 (fake).
    \item \textbf{Policy Update:} The policy $\pi_\theta$ is trained using a reinforcement learning algorithm, typically an on-policy method like Trust Region Policy Optimization (TRPO) \cite{schulman2017trustregionpolicyoptimization}. The key insight is that the discriminator's output provides a reward signal: $r(s, a) = -\log(1 - D_\phi(s, a))$ or $r(s,a) = \log(D_\phi(s, a))$. The policy is trained to maximize the expected return using this learned reward. As the policy improves, it produces $(s,a)$ pairs that are more successful at "fooling" the discriminator, thus receiving a higher reward.
\end{itemize}

\subsubsection*{Limitations of GAIL}
GAIL, particularly when using TRPO, has been shown to be stable and capable of learning complex policies in many environments. It also has the advantage of not requiring a large number of expert samples, unlike Behaviour Cloning. However, its primary drawback is \textbf{sample inefficiency}. This stems from its reliance on on-policy algorithms like TRPO, which limit the size of policy updates and discard samples after a single gradient step, thus requiring a large number of environment interactions to converge.

\section{Understanding Sample Inefficiency}

Sample inefficiency is a primary barrier to applying modern imitation learning methods to real-world problems. We analyze this challenge through the lens of GAIL, which, as previously mentioned, is a highly robust and principled method for imitation learning. Its main disadvantage, however, is its poor sample efficiency.

It is critical to distinguish between two forms of sample efficiency. Adversarial imitation learning methods like GAIL are often \textit{expert sample efficient}, requiring relatively few expert demonstrations. However, they are typically \textit{environment sample inefficient}, demanding a large number of agent-environment interactions to train the policy.

This inefficiency is problematic because extensive interaction can be costly, infeasible, or fundamentally alter the environment. For example, a physical robot may suffer wear and tear, or a financial market may shift during the long training time required. Hence, understanding and addressing the causes of environment sample inefficiency is essential for developing practical, robust imitation agents that can learn quickly.

\subsection{Causes of Sample Inefficiency}

The poor sample efficiency of GAIL can be attributed to several key factors:

\begin{itemize}
    \item \textbf{On-Policy Learning:} GAIL's policy optimization step typically uses an on-policy RL algorithm such as TRPO. By definition, on-policy methods are "data-hungry" because they discard all past trajectories after a single gradient update. The policy must continuously collect a new batch of trajectories from the environment for every incremental learning step, which is a principal source of inefficiency.

    \item \textbf{Instability from Interacting Networks:} The GAIL framework involves a complex, simultaneous optimization of at least three distinct parameterized functions: an actor ($\pi_\theta$, the policy), a critic ($V_\phi$, the state-value function), and a discriminator ($D_\omega$, the reward provider). The interaction between these networks can create instability, which in turn necessitates small, cautious policy updates (like those enforced by TRPO) to ensure convergence. This increases the total number of agent-environment interactions required.

    \item \textbf{Unbounded Stochastic Actions:} In continuous control, stochastic policies often output the parameters (e.g., mean and variance) of a Gaussian distribution. A Gaussian has infinite support, meaning it can sample actions in the range $(-\infty, +\infty)$. However, most real-world environments have hard, bounded action spaces (e.g., motor torque limits). Actions sampled outside this valid range are typically clipped. This clipping operation, however, provides no useful gradient signal to the policy to \textit{learn} the valid bounds. These "wasted" actions do not contribute to learning the optimal policy and only lead to an increased number of environment interactions (Chou and Scherer, 2017 \cite{pmlr-v70-chou17a}). 
\end{itemize}

\subsection{Methods to Improve Sample Efficiency}
\label{sec:methodsimprovesampleeffi}

Based on the causes identified, we propose several methods to improve the sample efficiency of adversarial imitation learning.

\begin{itemize}
    \item \textbf{Off-Policy Learning:}
    GAIL's sample inefficiency is fundamentally linked to its on-policy RL algorithm (e.g., TRPO). We propose replacing this with an \textbf{off-policy actor-critic algorithm}. The generator (policy) in the adversarial framework is not strictly dependent on on-policy updates. By using an off-policy method, we can leverage a \textbf{replay buffer} which stores transitions from both current and past policies. This allows for the re-use of environment interactions, significantly increasing the data available for gradient updates and leading to faster, more data-efficient convergence.

    \item \textbf{Reducing Network Complexity:}
    In the GAIL framework, the discriminator's primary role is to act as a learned reward provider for the policy improvement (RL) step. This introduces a complex three-way interaction (actor, critic, discriminator). We hypothesize that this complexity can be reduced. Since the discriminator is trained to assign high values (approaching 1) to expert pairs and low values (approaching 0) to non-expert pairs, this reward signal could potentially be incorporated directly into the value function learning, \textbf{removing the need for a separate discriminator network}. A reduction in the number of interacting parameterized functions is likely to reduce instability and, consequently, the overall sample complexity of the algorithm.

    \item \textbf{Bounded Stochastic Actions:}
    Stochastic policies based on unbounded distributions (e.g., Gaussian) are inefficient in bounded action spaces, as actions must be clipped, providing no useful gradient. We propose an actor architecture that \textbf{naturally limits its action outputs}. Inspired by DDPG ( \cite{lillicrap2019continuouscontroldeepreinforcement}), the actor network's final layer can use a $\tanh$ activation to bound its output to the range $[-1, +1]$. To maintain a stochastic policy, a noise vector $z \sim \mathcal{P}_z$ can be provided as an additional input to the network, along with the state $s_t$. The resulting $\tanh$-bounded output is then scaled and offset to match the environment's specific action bounds. This ensures all sampled actions are within the valid range, eliminating "wasted" samples and contributing to more efficient learning.

\end{itemize}

\section{Formulation of Algorithm Objectives}

\subsection{Off-Policy Actor-Critic}

The policy improvement step in GAIL typically employs an \textbf{on-policy} actor-critic structure. In this formulation, two parameterized functions are learned, both taking the state $s$ as input: an actor network that outputs the parameters of a stochastic policy $\pi_\theta(a|s)$, and a critic network that estimates the state-value function $V(s)$.

In contrast, an \textbf{off-policy} actor-critic, particularly for continuous action spaces, uses a different critic structure. While the actor function $\pi_\theta$ is similar, the critic learns the state-action value function $Q(s,a)$, taking both the state $s_t$ and the action $a_t$ as input.
\begin{itemize}
    \item For \textbf{continuous actions}, this off-policy structure is exemplified by the Deep Deterministic Policy Gradients (DDPG) algorithm \cite{lillicrap2019continuouscontroldeepreinforcement}.
    \item For \textbf{discrete actions}, the critic can estimate $Q(s,a)$ by taking only the state $s$ as input and outputting a value for each discrete action, similar to the architecture of a Deep Q-Network (DQN) \cite{mnih2013playingatarideepreinforcement}.
\end{itemize}

The general objective for the off-policy actor (policy) is to find parameters $\theta$ that maximize the expected Q-value of its actions:
\begin{equation}
    \max_{\theta} \quad \mathbb{E}_{s \sim \rho^\beta, a \sim \pi_\theta} [Q^{\pi_\theta, \nu}(s,a)]
    \label{eq:actor_objective}
\end{equation}
where $\rho^\beta$ is the state-visitation distribution of a behavior policy (or the replay buffer distribution) and $\nu$ are the parameters of the critic $Q$.

For a general \textbf{stochastic policy} $\pi_\theta(a|s)$, the policy gradient, as shown by \cite{degris2013offpolicyactorcritic}, is given by:
\begin{equation}
    \nabla_\theta J(\theta) = \mathbb{E}_{s \sim \rho^\beta, a \sim \pi_\theta} [ Q^{\pi_\theta, \nu}(s,a) \nabla_\theta \log \pi_\theta(a|s) ]
    \label{eq:stochastic_pg}
\end{equation}

However, as discussed in Section \ref{sec:methodsimprovesampleeffi} , we can use a reparameterized, \textbf{deterministic policy} $\pi_\theta(s, z)$ that takes noise $z$ as a separate input \cite{lillicrap2019continuouscontroldeepreinforcement}. This "deterministic" formulation (conditional on $z$) allows the gradients from the critic to flow directly to the actor, resulting in the following gradient update:
\begin{equation}
    \nabla_\theta J(\theta) = \mathbb{E}_{s \sim \rho^\beta, z \sim \mathcal{P}_z} [ \nabla_a Q^{\pi_\theta, \nu}(s,a) \big|_{a=\pi_\theta(s,z)} \nabla_\theta \pi_\theta(s,z) ]
    \label{eq:deterministic_pg}
\end{equation}

\begin{figure}[htbp]
    \centering
    \includegraphics[width=0.8\textwidth]{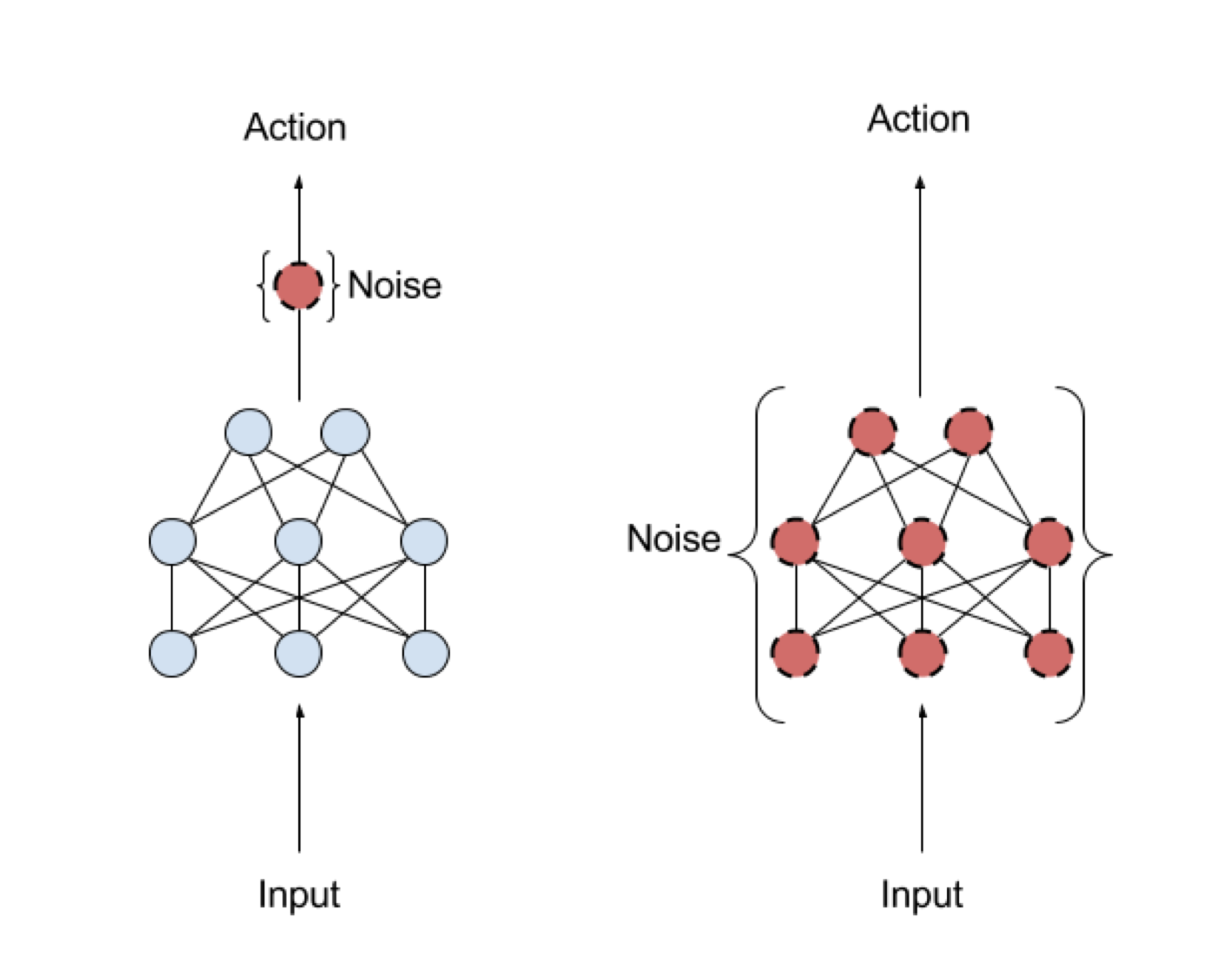}
    \caption{Using noise sampling after output vs Using noise as input and giving a bounded action output \cite{plappert2018parameterspacenoiseexploration}}
    \label{fig:network_junction}
\end{figure}

The primary advantage of this reparameterized actor is twofold. First, it preserves stochasticity for exploration by treating noise as an input to the network. Second, by applying a tanh activation to the actor's output layer, the action is naturally bounded. This output can then be scaled to match the environment's specific action limits, guaranteeing that all generated actions are feasible.

This approach ensures that actions are not "wasted" by clipping, thereby accelerating policy convergence and reducing sample complexity. By adopting the actor-critic structure from DDPG, we simultaneously achieve two goals: we transform the algorithm into an off-policy method capable of leveraging a replay buffer, and we enforce a bounded action policy.

\begin{figure}[htbp]
    \centering
    \includegraphics[width=0.8\textwidth]{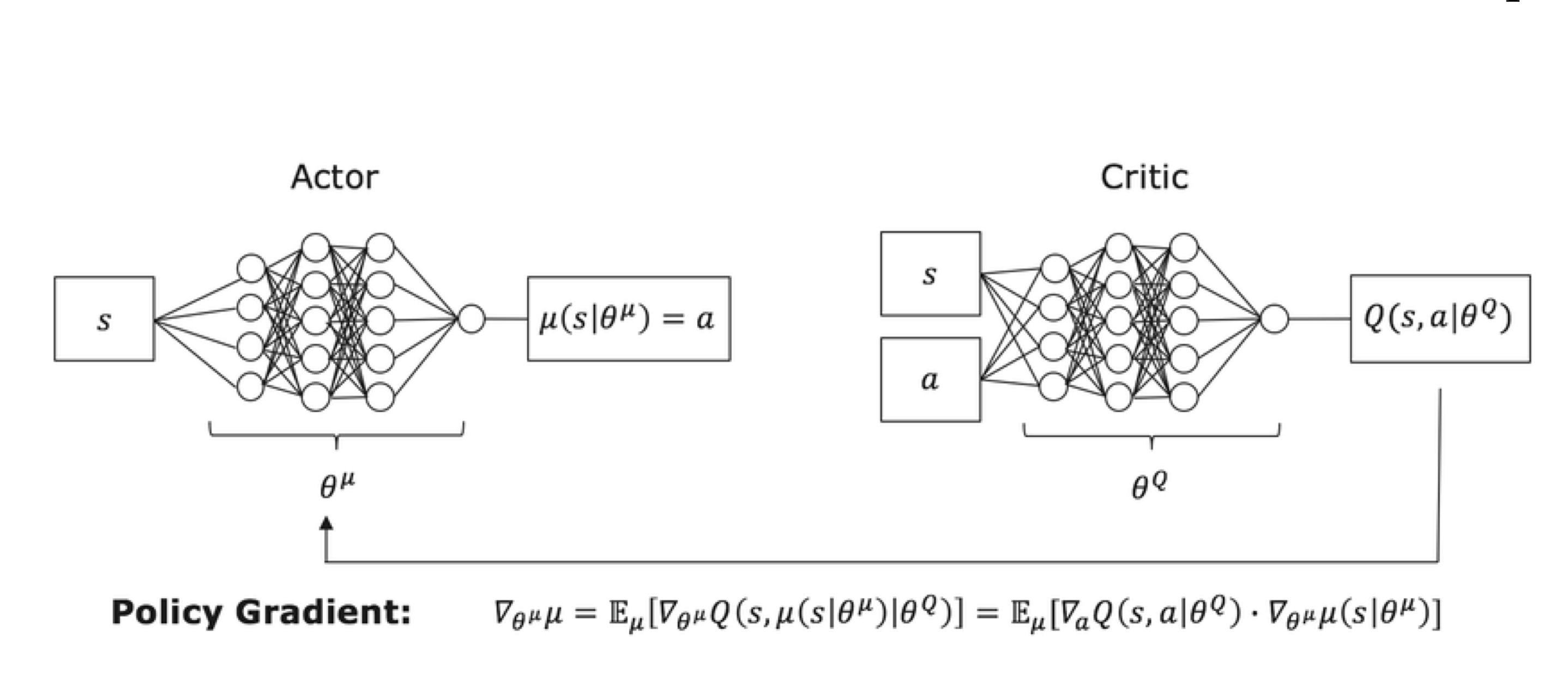}
    \caption{Oﬀ-policy actor critic architecture for continuous control \cite{liessnerdeeprlenergy2018}}
    \label{fig:offpolicyactorcritic}
\end{figure}

\subsection{Reward Learning}

We first define a reward function $R_\omega(s,a) = \log(r_\omega(s,a))$, where $r_\omega : \mathcal{S} \times \mathcal{A} \to [0, 1]$ is a function parameterized by $\omega$. This function $r_\omega$ is used to define a Bernoulli distribution $p_\omega : \Pi \times \mathcal{S} \times \mathcal{A} \to [0, 1]$ such that:
\begin{itemize}
    \item $p_\omega(\pi_E | s, a) = r_\omega(s, a)$ for the expert policy $\pi_E$, and
    \item $p_\omega(\pi | s, a) = 1 - r_\omega(s, a)$ for all other policies $\pi \in \Pi \setminus \{\pi_E\}$, which include our learned policy $\pi_\theta$ and the behavior policy $\beta$ from the replay buffer.
\end{itemize}

The original max-margin IRL objective can be expressed as:
\begin{equation}
    \text{IRL}(\pi_E) = \argmax_{\omega} \quad \left[ J(\pi_E, R_\omega) - J(\pi, R_\omega) \right]
    \label{eq:irl_objective}
\end{equation}
To ensure \ref{eq:irl_objective} is $\ge 0$, a simple reward definition would be to assign $r_\omega = 1$ for expert state-action pairs $(s, a) \in \mathcal{S}_E \times \mathcal{A}_E$ and $r_\omega = 0$ for non-expert pairs $(s, a) \in \mathcal{S}_\pi \times \mathcal{A}_\pi$.

However, using this binary definition with our log-space reward $R_\omega = \log(r_\omega)$ would cause the return for non-expert policies to become $-\infty$ (since $\log(0) = -\infty$), leading to numerical instability.

Therefore, we define our IRL objective in a different way, optimizing for a combination of expert likelihood and non-expert entropy:
\begin{equation}
    \argmax_{\omega} \quad \mathbb{E}_{\substack{s \sim \rho_{\pi_E} \\ a \sim \pi_E}} [p_\omega(\pi_E | s, a)] + \mathbb{E}_{\substack{s \sim \rho_\pi \\ a \sim \pi}} [H(p_\omega(\cdot | s, a))]
    \label{eq:new_irl_objective}
\end{equation}
where $H$ denotes the entropy of the Bernoulli distribution $p_\omega$:
\begin{equation}
\begin{split}
    H(p_\omega(\cdot | s, a)) = & -p_\omega(\pi_E | s, a) \log p_\omega(\pi_E | s, a) \\
                             & -p_\omega(\pi | s, a) \log p_\omega(\pi | s, a)
\end{split}
\label{eq:bernoulli_entropy}
\end{equation}

This objective has a desirable property: for non-expert pairs $(s, a) \in \mathcal{S}_\pi \times \mathcal{A}_\pi$, the optimal solution for the second term is to assign maximum uncertainty, i.e., $p_\omega(\pi_E | s, a) = p_\omega(\pi | s, a) = 0.5$. This assignment explicitly maximizes the entropy, which in turn makes the return for the non-expert policy finite and stable. This formulation satisfies the max-margin constraint of Equation~\ref{eq:irl_objective} without explicitly discriminating between expert and non-expert pairs with unstable reward values.

\subsection{Value Learning}

Given that the reward function has been modeled as a log-probability, the discounted return can also be modeled as a log-probability. It follows naturally that the $Q$-value function should also be represented in this form. The  function can be written as \cite{sasaki2018sample} $q_{\pi_\theta, \nu} : \mathcal{S} \times \mathcal{A} \to [0, 1]$, parameterized by $\nu$, to serve as a probabilistic approximator for the $Q$-function:
\begin{equation}
    Q_{\pi_\theta, \nu}(s_t, a_t) = \log(q_{\pi_\theta, \nu}(s_t, a_t))
    \label{eq:q_log_prob}
\end{equation}
Recall that the standard Bellman equation for a policy $\pi$ is written as:
\begin{equation}
    Q_\pi(s_t, a_t) = R(s_t, a_t) + \gamma \mathbb{E}_{s_{t+1} \sim \tau, a_{t+1} \sim \pi} [Q_\pi(s_{t+1}, a_{t+1})]
    \label{eq:standard_bellman}
\end{equation}
We now substitute our newly defined terms into this equation, replacing $\pi$ with $\pi_\theta$, $R(s_t, a_t)$ with $\log(r_\omega(s_t, a_t))$, and $Q_{\pi_\theta, \nu}$ with $\log(q_{\pi_\theta, \nu})$. This yields a modified Bellman equation in the log-probability domain:
\begin{equation}
    \log(q_{\pi_\theta, \nu}(s_t, a_t)) = \mathbb{E}_{\substack{s_{t+1} \sim \tau \\ a_{t+1} \sim \pi}} \left[ \log \left( r_\omega(s_t, a_t) \cdot q_{\pi_\theta, \nu}^\gamma(s_{t+1}, a_{t+1}) \right) \right]
    \label{eq:modified_bellman}
\end{equation}
To formalize this probabilistic structure, we introduce additional random variables. First, a policy-conditioned probability given the current state-action pair:
\begin{equation}
    P_\nu(\pi | s_t, a_t) = 
    \begin{cases} 
        q_{\pi_\theta, \nu}(s, a), & \text{if } \pi = \pi_E \\
        1 - P_\nu(\pi_E | s_t, a_t), & \text{otherwise}
    \end{cases}
    \label{eq:prob_pi_given_sa}
\end{equation}
Second, a transition-and-policy-conditioned probability:
\begin{equation}
    P_{\omega, \nu, \gamma}(\pi | s_t, a_t, s_{t+1}, a_{t+1}) = 
    \begin{cases} 
        r_\omega(s_t, a_t) \cdot q_{\pi_\theta, \nu}^\gamma(s_{t+1}, a_{t+1}), & \text{if } \pi = \pi_E \\
        1 - P_{\omega, \nu, \gamma}(\pi_E | s_t, a_t, s_{t+1}, a_{t+1}), & \text{otherwise}
    \end{cases}
    \label{eq:prob_pi_given_transition}
\end{equation}

Using these probabilistic definitions, the Bellman error, which we aim to minimize, can be defined as the loss $L(\omega, \nu, \theta)$. This loss is the difference between the two sides of our modified Bellman equation (Equation~\ref{eq:modified_bellman}):
\begin{equation}
\begin{split}
    L(\omega, \nu, \theta) = & \E_{\substack{s_{t+1} \sim \tau \\ a_{t+1} \sim \pi_\theta}} \left[ \log P_{\omega, \nu, \gamma}(\pi_E | s_t, a_t, s_{t+1}, a_{t+1}) \right] \\
                           & - \log P_\nu(\pi_E | s_t, a_t)
\end{split}
\label{eq:loss_L}
\end{equation}
By applying Jensen's inequality to the concave $\log$ function, we can establish an upper bound on this loss:
\begin{equation}
    L(\omega, \nu, \theta) \le \log \left( \frac{\E_{\substack{s_{t+1} \sim \tau \\ a_{t+1} \sim \pi_\theta}} [P_{\omega, \nu, \gamma}(\pi_E | \dots)]}{P_\nu(\pi_E | s_t, a_t)} \right)
\label{eq:jensen_bound}
\end{equation}
The loss $L(\omega, \nu, \theta)$ is effectively measuring the ratio between two Bernoulli distributions, $P_\nu$ and the expected target distribution $\E[P_{\omega, \nu, \gamma}]$. The loss $L=0$ only when these two distributions are matched:
\begin{equation*}
    \E_{\substack{s_{t+1} \sim \tau \\ a_{t+1} \sim \pi_\theta}} [P_{\omega, \nu, \gamma}(\pi_E | \dots)] = P_\nu(\pi_E | s_t, a_t)
\end{equation*}
Since the objective is to match these two distributions, we can reframe the value learning problem as minimizing the divergence between them. We use the Jensen-Shannon divergence ($\DJS$) as a stable metric. Thus, the final objective for optimizing the critic $q_{\pi_\theta, \nu}$ (parameterized by $\nu$) is:
\begin{equation}
    \argmin_{\nu} \quad \E \left[ \DJS \left( P_\nu(\cdot | s_t, a_t) \parallel \E_{\substack{s_{t+1} \sim \tau \\ a_{t+1} \sim \pi_\theta}} [P_{\omega, \nu, \gamma}(\cdot | s_t, a_t, s_{t+1}, a_{t+1})] \right) \right]
    \label{eq:djs_objective}
\end{equation}
where the outer expectation is over the state-action pairs $(s_t, a_t)$ sampled from the replay buffer.

\subsection{Incorporating Reward Learning into the Critic}

As previously discussed, reducing the number of simultaneously optimized networks can improve stability. We can merge the reward learning step \cite{sasaki2018sample} directly into the critic's update, thereby eliminating the separate discriminator network parameterized by $\omega$.

We first assume access to the \textit{optimal} reward function $R^*_\omega(s,a) = \log r^*_\omega(s,a)$ derived from our objective in Equation~\ref{eq:new_irl_objective}. This optimal reward function implicitly assigns:
\begin{itemize}
    \item $r^*_\omega(s_t, a_t) = 1$ (and $\log r^*_\omega = 0$) for expert pairs $(s_t, a_t) \in \mathcal{S}_E \times \mathcal{A}_E$.
    \item $r^*_\omega(s_t, a_t) = 0.5$ (and $\log r^*_\omega = \log(0.5) = -\log 2$) for non-expert pairs $(s_t, a_t) \in \mathcal{S}_\pi \times \mathcal{A}_\pi$.
\end{itemize}

We can rearrange the modified Bellman equation (Equation~\ref{eq:modified_bellman}) to isolate the reward term, which defines the target for our Bellman error:
\begin{equation}
    \log r^*_\omega(s_t, a_t) = \log q_{\pi_\theta, \nu}(s_t, a_t) - \E_{\substack{s_{t+1} \sim \tau \\ a_{t+1} \sim \pi_\theta}} \left[ \log(q_{\pi_\theta, \nu}^\gamma(s_{t+1}, a_{t+1})) \right]
    \label{eq:isolated_reward}
\end{equation}
By substituting the known optimal values for $r^*_\omega$ into this equation, we can formulate a new objective for the critic $\nu$. This objective is the sum of two loss terms, one over expert data ($\mathcal{D}_E$) and one over policy/buffer data ($\mathcal{D}_\pi$):
\begin{equation}
\begin{split}
    \argmin_{\nu} \quad & \E_{\substack{s_t \sim \rho_{\pi_E} \\ a_t \sim \pi_E}} \left[ \log q_{\pi_\theta, \nu}(s_t, a_t) - \E_{\substack{s_{t+1} \sim \tau \\ a_{t+1} \sim \pi_\theta}}[\log(q_{\pi_\theta, \nu}^\gamma(s_{t+1}, a_{t+1}))] \right] \\
    + & \E_{\substack{s_t \sim \rho_\pi \\ a_t \sim \pi}} \left[ \log q_{\pi_\theta, \nu}(s_t, a_t) - \E_{\substack{s_{t+1} \sim \tau \\ a_{t+1} \sim \pi_\theta}}[\log(q_{\pi_\theta, \nu}^\gamma(s_{t+1}, a_{t+1}) / 2)] \right]
\end{split}
\label{eq:combined_critic_loss}
\end{equation}
The first term minimizes the Bellman error toward its target of $0$ (since $\log 1 = 0$). The second term minimizes the Bellman error toward its target of $\log(0.5)$ (which is $-\log 2$, correctly represented by the division by 2 inside the logarithm).

This objective (Equation~\ref{eq:combined_critic_loss}) is the sum of two Bellman error terms, analogous to the loss defined in Equation~\ref{eq:loss_L}. Following the same logic that transformed Equation~\ref{eq:loss_L} into the divergence minimization objective (Equation~\ref{eq:djs_objective}), we can express our new critic loss as the sum of two Jensen-Shannon divergences:
\begin{equation}
\begin{split}
    \argmin_{\nu} \quad & \E_{\substack{s_t \sim \rho_{\pi_E} \\ a_t \sim \pi_E}} \left[ \DJS \left( P_\nu(\cdot | s_t, a_t) \parallel \E_{\substack{s_{t+1} \sim \tau \\ a_{t+1} \sim \pi_\theta}} [P_\nu^\gamma(\cdot | s_{t+1}, a_{t+1})] \right) \right] \\
    + & \E_{\substack{s_t \sim \rho_\pi \\ a_t \sim \pi}} \left[ \DJS \left( P_\nu(\cdot | s_t, a_t) \parallel \E_{\substack{s_{t+1} \sim \tau \\ a_{t+1} \sim \pi_\theta}} [P_\nu^\gamma(\cdot | s_{t+1}, a_{t+1}) / 2] \right) \right]
\end{split}
\label{eq:combined_djs_loss}
\end{equation}
where $P_\nu^\gamma(\dots)$ is shorthand for $q_{\pi_\theta, \nu}^\gamma(\dots)$ and represents the target distribution (implying $r=1$), and $P_\nu^\gamma(\dots) / 2$ represents the target distribution (implying $r=0.5$).

Critically, the optimization objective in Equation~\ref{eq:combined_djs_loss} is no longer dependent on the parameters $\omega$. By embedding the optimal reward values directly into the critic's target distributions, the critic update step now implicitly learns the reward function. This removes the need for a separate discriminator network, reducing the algorithm's complexity and potential for instability.

\section{Implementation Issues and Stabilization}

A primary challenge for all off-policy RL algorithms, including ours, is training stability. It is well-known that Q-learning with non-linear function approximators (i.e., deep neural networks) and the standard temporal difference (TD) error is not guaranteed to converge. To address this, we incorporate two key stabilization techniques from modern deep RL.

\begin{enumerate}
    \item \textbf{Soft Target Updates:}
    Standard Q-learning involves chasing a "moving target," as the target Q-value $Q(s', a')$ is generated by the same network $Q_\nu$ whose parameters $\nu$ are being actively updated. This can lead to oscillations and divergence. To provide a stable target, we use a separate target network $Q_{\nu_{\text{target}}}$. The main critic network $Q_\nu$ is updated via gradient descent, but the target network $Q_{\nu_{\text{target}}}$ is updated slowly using a "soft" update  \cite{lillicrap2019continuouscontroldeepreinforcement}. After each gradient step, the target network parameters are updated as a moving average:
    \begin{equation}
        \nu_{\text{target}} \leftarrow \tau \nu + (1 - \tau) \nu_{\text{target}}
        \label{eq:soft_update}
    \end{equation}
    where $\tau \ll 1$ (e.g., $\tau = 0.001$) is a hyperparameter that controls the update speed.
    \item \textbf{Clipped Double Q-Learning (TD3):}
    To combat the overestimation bias inherent in Q-learning, we adopt the "Clipped Double Q-learning" technique from TD3  \cite{fujimoto2018addressingfunctionapproximationerror}. This method maintains \textit{two} independent critic networks, $Q_{\nu_1}$ and $Q_{\nu_2}$, along with their corresponding target networks, $Q_{\nu_{1, \text{target}}}$ and $Q_{\nu_{2, \text{target}}}$.
    When computing the target value for the Bellman update, the \textit{minimum} of the two target networks is used:
    \begin{equation}
        Q_{\text{target}}(s', a') = \min(Q_{\nu_{1, \text{target}}}(s', a'), Q_{\nu_{2, \text{target}}}(s', a'))
    \end{equation}
    Both critic networks are then trained to regress to this more conservative target. Our JSD-based loss from Equation~\ref{eq:combined_djs_loss} is thus modified to be the sum of the losses for each critic, both using this common target:
    \begin{equation}
        L_{\text{critic}} = L_{\DJS}(Q_{\nu_1}, Q_{\text{target}}) + L_{\DJS}(Q_{\nu_2}, Q_{\text{target}})
        \label{eq:td3_loss}
    \end{equation}
    Both target networks are then updated softly using their respective main networks:
    \begin{align}
        \nu_{1, \text{target}} & \leftarrow \tau \nu_1 + (1 - \tau) \nu_{1, \text{target}} \label{eq:td3_soft_1} \\
        \nu_{2, \text{target}} & \leftarrow \tau \nu_2 + (1 - \tau) \nu_{2, \text{target}} \label{eq:td3_soft_2}
    \end{align}
    This prevents a single network from developing an overly optimistic estimate of the Q-value, which significantly enhances stability.
\end{enumerate}

\section{Algorithm}
The complete off-policy imitation learning procedure is detailed in Algorithm~\ref{alg:main} enhancing the method shown in \cite{sasaki2018sample}. The algorithm maintains two replay buffers: a static buffer $\BE$ for expert demonstrations and a replay buffer $\BBeta$ for transitions collected by the agent. It adopts a "Clipped Double Q-learning" (TD3) style architecture with two critics ($\nu_1, \nu_2$) and their corresponding target networks ($\nu_{1,\text{target}}, \nu_{2,\text{target}}$) to stabilize training. The critic networks are updated using the combined Jensen-Shannon Divergence loss derived in Equation~\ref{eq:combined_djs_loss}, where the target distribution is formed using the minimum of the two target Q-networks. The actor is updated using the policy gradient from Equation~\ref{eq:deterministic_pg}.

\begin{algorithm}[htbp]
\small
\caption{Off-Policy Imitation Learning}
\label{alg:main}
\SetAlgoLined
Initialize actor $\theta$, critic networks $\nu_1, \nu_2$\;
Initialize target networks $\nu_{1,\text{target}} \leftarrow \nu_1$, $\nu_{2,\text{target}} \leftarrow \nu_2$\;
Initialize replay buffer $\BBeta \leftarrow \emptyset$\;
Load expert trajectories into static buffer $\BE$\;
 
\For{ep = 1 to M}{
    $t \leftarrow 0$\;
    Reset environment, receive initial state $s_0$\;
    
    \While{not done}{
        Sample action $a_t \sim \pi_\theta(\cdot | s_t, z_t)$ \tcp{Stochastic action from noise-augmented policy}
        Execute $a_t$, observe next state $s_{t+1}$ and `done` flag\;
        Store transition $(s_t, a_t, s_{t+1}, \text{done})$ in $\BBeta$\;
        $s_t \leftarrow s_{t+1}$\;
        $t \leftarrow t + 1$\;
    }
    
    \For{u = 1 to t}{
        \tcp{Perform $t$ gradient updates}
        Sample mini-batch $\mathcal{B}_E = \{(s^E, a^E, s'^E)\}$ from $\BE$\;
        Sample mini-batch $\mathcal{B}_\beta = \{(s^\beta, a^\beta, s'^\beta, d^\beta)\}$ from $\BBeta$\;
        Let $\mathcal{B}_{s'} = \{s'^E\} \cup \{s'^\beta\}$ be the set of all next states\;
        
        \tcp{--- Critic Update ---}
        Compute target actions for next states: $a' \sim \pi_\theta(\cdot | s', z')$ for $s' \in \mathcal{B}_{s'}$\;
        Compute target Q-value (using $\log(q) = Q$): 
        $$Q'_{\text{min}} \leftarrow \min_{i=1,2} \log( q_{\pi_\theta, \nu_{i, \text{target}}}(s', a') )$$\;
        Convert to target distribution: $\Ptarget \leftarrow \exp(Q'_{\text{min}})$\;
        Update $\nu_1, \nu_2$ by minimizing the sum of JSD losses (from Eq.~\ref{eq:combined_djs_loss}):
        $$L \leftarrow \sum_{i=1,2} \left( \E_{\mathcal{B}_E} [\DJS( P_{\nu_i} \parallel \E[\Ptarget] )] + \E_{\mathcal{B}_\beta} [\DJS( P_{\nu_i} \parallel \E[\Ptarget / 2] )] \right)$$\;
        
        \tcp{--- Actor Update ---}
        Sample mini-batch $\mathcal{B}'_\beta = \{(s, a, s', d)\}$ from $\BBeta$\;
        Update $\theta$ by policy gradient (from Eq.~\ref{eq:deterministic_pg}, using $\nu_1$):
        $$\nabla_\theta J(\theta) \approx \E_{s \sim \mathcal{B}'_\beta} [ \nabla_a \log( q_{\pi_\theta, \nu_1}(s,a) ) \big|_{a=\pi_\theta(s,z)} \nabla_\theta \pi_\theta(s,z) ]$$\;
        
        \tcp{--- Target Network Soft Update ---}
        $\nu_{i, \text{target}} \leftarrow \tau \nu_i + (1-\tau)\nu_{i, \text{target}}$ for $i=1,2$\;
    }
}
\end{algorithm}

\section{Experiments}

We evaluate our algorithm in the OpenAI Gym suite \citep{brockman2016openai}, a standard benchmarking platform for reinforcement learning.

\subsection{Environment}

We select the \texttt{BipedalWalker-v2} environment, a challenging continuous control task. The objective is for a bipedal robot to learn to walk across uneven terrain.
\begin{itemize}
    \item \textbf{State Space:} The environment has a 24-dimensional continuous state space ($\mathcal{S} \subseteq \mathbb{R}^{24}$), including hull angles, joint positions, and LiDAR sensor readings.
    \item \textbf{Action Space:} The action space is 4-dimensional and continuous ($\mathcal{A} \subseteq [-1, 1]^4$), representing the torque applied to the hip and knee joints of both legs.
    \item \textbf{Reward Structure:} The agent receives a positive reward proportional to the forward distance traversed. A large negative penalty of -100 is applied if the agent falls, which also terminates the episode. Additional positive rewards are provided for moving quickly. The maximum episodic return for an expert policy is approximately 330.
\end{itemize}

\begin{figure}[htbp]
    \centering
    \includegraphics[width=0.8\textwidth]{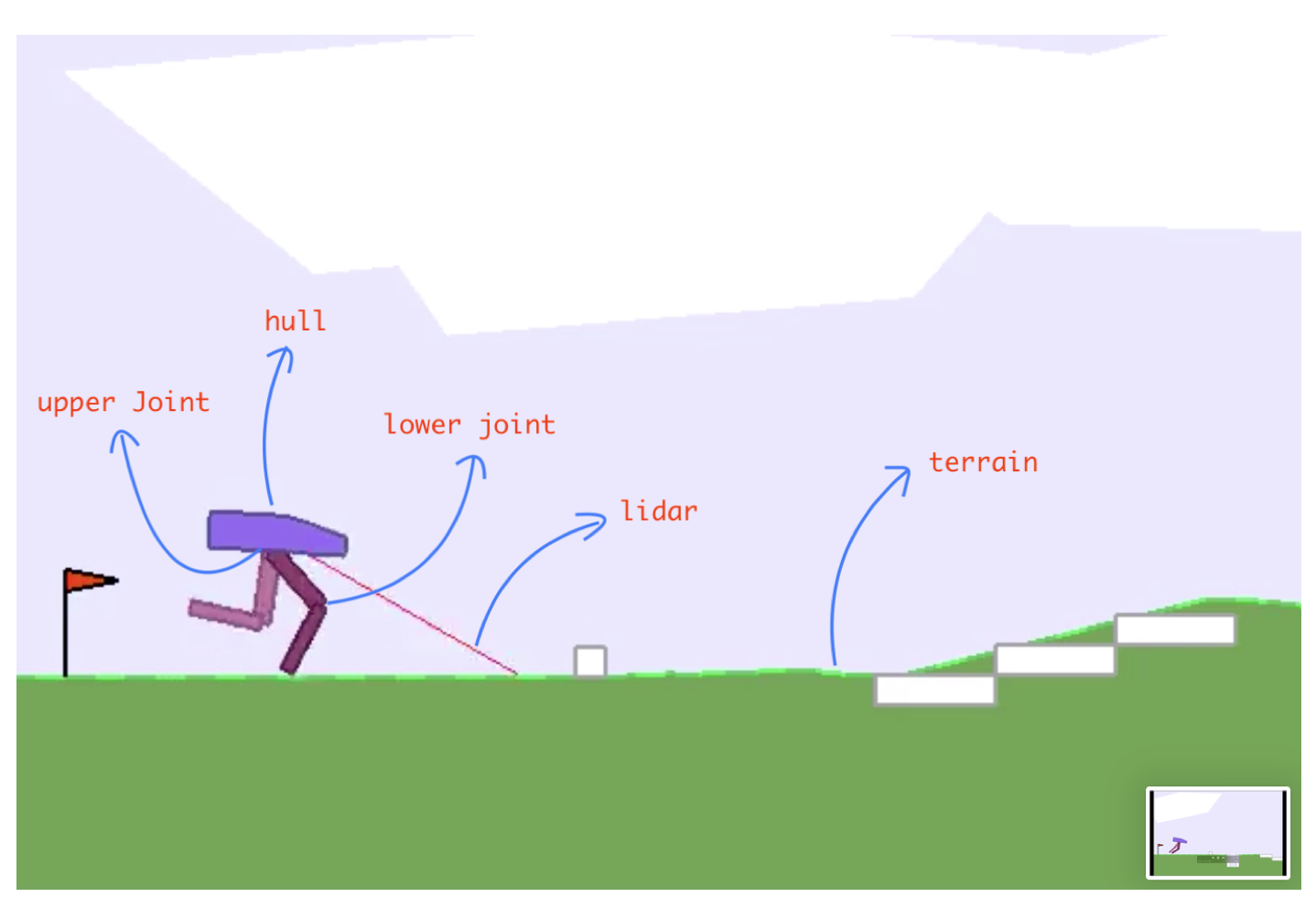}
    \caption{BipedalWalker-v2 environment in OpenAI gym}
    \label{fig:bipedalwalkerv2}
\end{figure}

\subsection{Expert Data Generation}

To create a dataset of expert demonstrations, we first trained a near-optimal policy using Proximal Policy Optimization (PPO) \citep{schulman2017proximalpolicyoptimizationalgorithms}, leveraging the implementation from the stable-baselines repository \citep{stable-baselines}. The PPO agent was trained for 5 million environment steps.

From this fully trained PPO policy, we collected a large set of trajectories. We then filtered this set, retaining only those trajectories where the total episodic return was greater than 300. This process yielded our final expert dataset, $\mathcal{D}_E$, consisting of 100 high-performance trajectories.

\section{Results and Conclusion}

\begin{figure}[htbp]
    \centering
    % Make sure the file 'Graph_Imitation.jpg' is uploaded to your Overleaf project
    \includegraphics[width=0.9\linewidth]{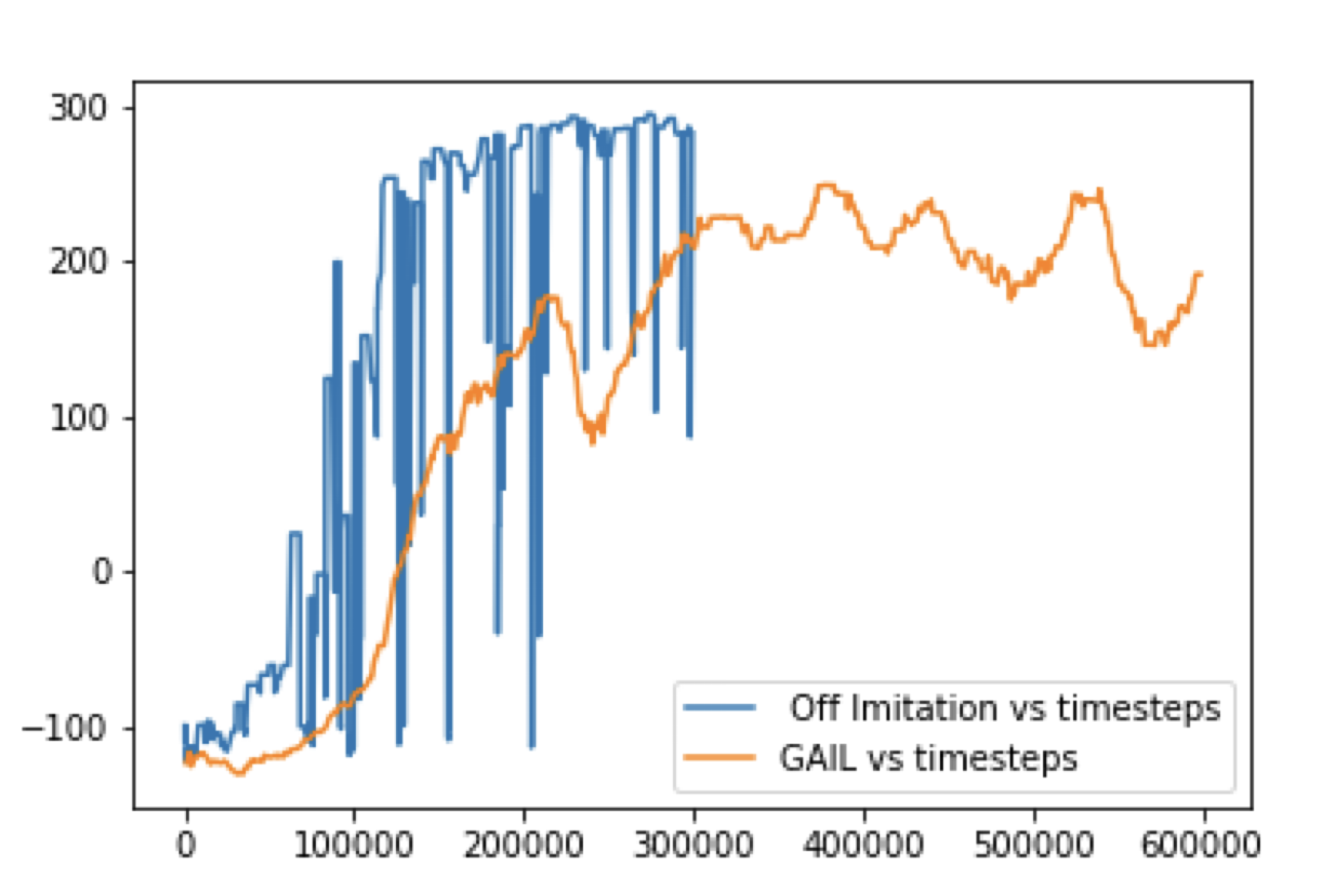} 
    \caption{Training performance of our off-policy imitation algorithm (blue) compared to the GAIL baseline (orange) on \texttt{BipedalWalker-v2}. Our method achieves expert-level rewards (~300) significantly faster, demonstrating a substantial improvement in sample efficiency.}
    \label{fig:resultsimitationoffpolicy}
\end{figure}

We evaluated our proposed off-policy imitation learning algorithm against the GAIL baseline on the \texttt{BipedalWalker-v2} environment. The results, shown in Figure~\ref{fig:resultsimitationoffpolicy}, demonstrate a clear and significant improvement in sample efficiency.

Our method (labeled "Off Imitation") learns remarkably quickly, achieving expert-level episodic returns of approximately 300 within only 200,000 environment timesteps. In contrast, the on-policy GAIL baseline learns much more slowly and plateaus at a lower performance level.

Critically, our algorithm achieves this high-performance policy \textbf{without access to the environment's true reward function}, relying solely on the provided expert trajectories. On training, our imitation algorithm manages to learn a policy that reaches rewards surpassing the GAIL baseline and matching the PPO-trained expert from which the demonstrations were collected.

This work successfully demonstrates that by combining an off-policy architecture, a bounded actor, and a stable JSD-based critic objective, we can drastically reduce the sample complexity of adversarial imitation learning. This makes the approach far more viable for complex, real-world tasks where environment interactions are expensive.

\section{Future Work}

While our results demonstrate a significant improvement in sample efficiency, the high variance in policy returns during training (Figure~\ref{fig:resultsimitationoffpolicy}) highlights a common challenge with off-policy methods. Our future work will focus on two primary directions to address this and extend the algorithm's applicability:

\begin{itemize}
    \item \textbf{Hybrid On-Policy/Off-Policy Updates:}
    To reduce variance and improve stability, we plan to explore hybrid algorithms. On-policy methods are known for their stable learning dynamics, while off-policy methods provide sample efficiency. A promising avenue is to combine these strengths, for instance, by using on-policy updates for the actor while retaining the off-policy, buffer-based update for the critic, as suggested by \cite{gu2017qpropsampleefficientpolicygradient}.

    \item \textbf{Learning from Suboptimal Demonstrations:}
    Our current framework assumes that the provided demonstrations are (near-)optimal. An important and orthogonal extension is to incorporate rewards from the environment alongside the expert demonstrations. This would allow the agent to learn effectively even from suboptimal data, as the environment's reward signal would provide an objective ground truth, enabling the policy to surpass the quality of the demonstrations. We aim to take help from \cite{fan2024imitationlearningsuboptimaldemonstrations} to further enhance our algorithm while accounting for stability.
\end{itemize}

\bibliographystyle{plain}
\bibliography{references}

%%%%%%%%%%%%%%%%%%%%%%%%%%%%%%%%%%%%%%%%%%%%%%%%%%%%%%%%%%%%

% \appendix

% \section{Appendix}

\end{document}